\def\AIPerfArxiv{1}
\ifdefined\AIPerfAnonymous\else
\def\AIPerfArxiv{1}
\fi
\RequirePackage[T1]{fontenc}
\documentclass[10pt,conference]{IEEEtran}
\usepackage{cite}
\usepackage{amsmath,amssymb}
\usepackage{graphicx,booktabs,array,multirow,tabularx,hhline}
\usepackage{xcolor,colortbl}
\usepackage{tcolorbox}
\usepackage[hidelinks]{hyperref}

\hypersetup{pdftitle={General-Purpose Decision Models for AI Configuration Tuning},pdfauthor={}}
\ifdefined\AIPerfArxiv
\hypersetup{pdfauthor={Jingzhi Gong and Jie M. Zhang}}
\fi

\definecolor{tablequality}{HTML}{4472C4}
\definecolor{tablecost}{HTML}{ED7D31}
\definecolor{tableruntime}{HTML}{70AD47}
\colorlet{tabletraining}{tableruntime}
\colorlet{tableerror}{tablequality}
\colorlet{tablelatency}{tableruntime}
\colorlet{tableperplexity}{tablequality}
\definecolor{tablecalls}{HTML}{BBBBBB}
\definecolor{tableheader}{HTML}{EDF0F3}
\newcommand{\metriccell}[3]{\cellcolor{#1!#2}#3}
\newcommand{\jevmodel}{\texorpdfstring{\texttt{Jev1.14.0}}{Jev1.14.0}}
\newcommand{\deepseekmodel}{\texorpdfstring{\texttt{DeepSeek-V4-Flash}}{DeepSeek-V4-Flash}}
\newcommand{\jev}{\texorpdfstring{\texttt{Jev}}{Jev}}
\newcommand{\deepseek}{\texorpdfstring{\texttt{DS-v4-F}}{DS-v4-F}}
\newcommand{\randommethod}{\texorpdfstring{\texttt{Random}}{Random}}
\newcommand{\flash}{\texorpdfstring{\texttt{FLASH}}{FLASH}}
\newcommand{\smac}{\texorpdfstring{\texttt{SMAC}}{SMAC}}
\newcommand{\hebo}{\texorpdfstring{\texttt{HEBO}}{HEBO}}
\newcommand{\promisetune}{\texorpdfstring{\texttt{PromiseTune}}{PromiseTune}}
\newcommand{\rf}{\texorpdfstring{\texttt{RF}}{RF}}
\newcommand{\gp}{\texorpdfstring{\texttt{GP}}{GP}}
\newcommand{\dal}{\texorpdfstring{\texttt{DaL}}{DaL}}
\newcommand{\llmperf}{\texorpdfstring{\texttt{LLM4Perf}}{LLM4Perf}}
\newcommand{\noregion}{\texorpdfstring{\texttt{No guider}}{No guider}}
\newcommand{\treeregion}{\texorpdfstring{\texttt{DT guider}}{DT guider}}

\newtcolorbox{rqanswer}{colback=tablequality!8,colframe=tablequality!8,
boxrule=0pt,arc=2pt,left=5pt,right=5pt,top=3pt,bottom=3pt,
before skip=6pt,after skip=6pt}
\newtcolorbox{implicationbox}{colback=tabletraining!10,colframe=tabletraining!45,
boxrule=0.4pt,arc=2pt,left=5pt,right=5pt,top=4pt,bottom=4pt,
before skip=6pt,after skip=6pt}
\title{General-Purpose Decision Models for AI Configuration Tuning}
\ifdefined\AIPerfArxiv
\author{\IEEEauthorblockN{Jingzhi Gong and Jie M. Zhang}
\IEEEauthorblockA{King's College London, United Kingdom\quad
\{jingzhi.gong, jie.zhang\}@kcl.ac.uk}}
\IEEEoverridecommandlockouts
\IEEEaftertitletext{\vspace{-12pt}}
\else
\author{\IEEEauthorblockN{Anonymous Authors}}
\fi
\begin{document}
\maketitle
\begin{abstract}
AI configuration tuning searches for settings that improve the performance of AI systems under a limited budget. Existing tuners use different sampling and search policies, making the choice of decision strategy central to both tuning quality and cost. We investigate whether general-purpose decision models can guide tuning, using \jev{} as a deployed representative that returns structured choices and numeric scores across diverse tasks. To assess where it helps, we compare its use as an optimizer, performance predictor, and region guider against conventional methods and a large language model (LLM) on eight AI tuning tasks. Conventional optimizers attain rank 1 in six of eight comparisons. Similarly, established predictors outperform \jev{} in seven. Through region guidance, however, it finds configurations with lower perplexity than those selected by the native tuner, offering a quality gain when additional decision time and API fees are acceptable. These findings yield actionable recommendations on selecting the decision role, retaining conventional predictors, and balancing tuning quality against cost.
\end{abstract}

\section{Introduction}
AI systems expose configuration choices that affect both computational efficiency and predictive quality. Neural-network training depends on settings such as batch size and learning rate~\cite{lcbench}, while language-model architecture affects perplexity and hardware-dependent inference latency~\cite{hwgpt}. Evaluating these choices consumes resources, making the available measurement budget central to configuration selection~\cite{aspirations}. Automated tuning therefore seeks useful configurations within that budget.

\begin{figure*}[t]
\centering
\includegraphics[width=\textwidth]{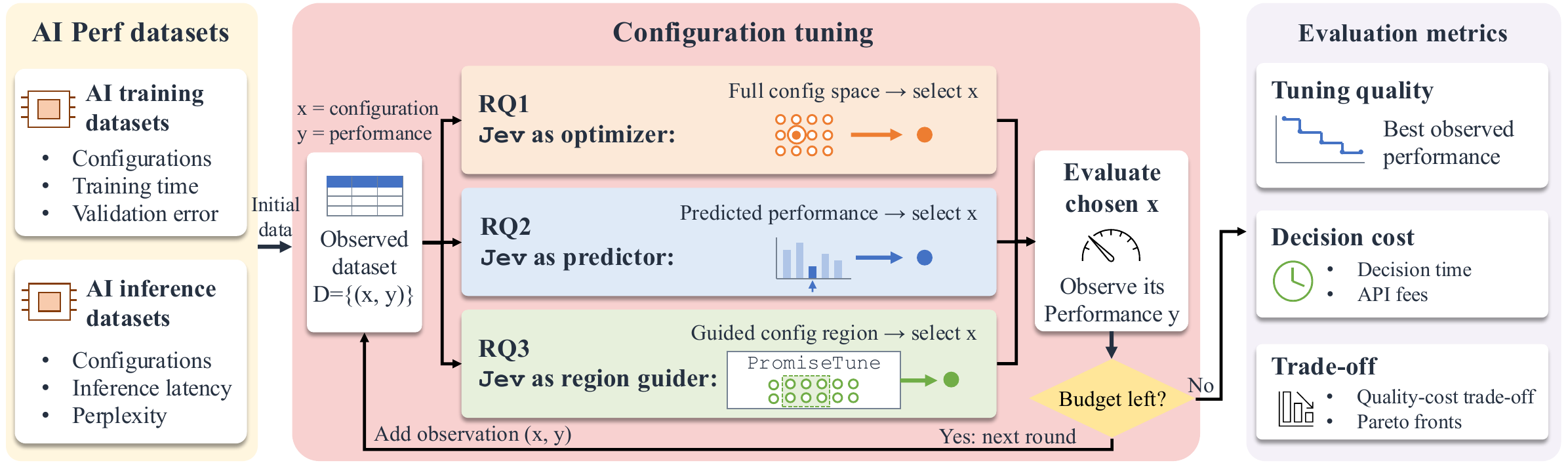}
\caption{Sequential configuration tuning through three decision roles, evaluated by attained performance, decision cost, and their trade-off.}
\label{fig:workflow}
\end{figure*}

Established tuners implement different decision strategies. \randommethod{}~\cite{random} samples configurations independently of their measured performance, while \flash{}~\cite{flash}, \smac{}~\cite{smac}, and \hebo{}~\cite{hebojournal} use learned predictors to guide selection. \promisetune{}~\cite{promisetune} additionally recommends promising regions of the configuration space. These strategies expose different opportunities for a decision model to contribute, from proposing the next configuration to guiding one component.

We investigate these opportunities using \jev{}, a deployed general-purpose decision model that returns structured choices and numeric scores for diverse tasks~\cite{jevintro,jevscore}. Studies of its use in agent workflows report mixed effects on success and total cost~\cite{jevworkflow}. Configuration tuning raises a related question because each model output can affect both the next evaluation and the evidence available for subsequent choices. We examine how this sequence changes when \jev{} selects configurations directly or guides an existing tuner.

We test these alternatives through the three roles in Figure~\ref{fig:workflow}. The optimizer proposes the next configuration, the predictor estimates candidate performance, and the region guider recommends a search region~\cite{flash,smac,promisetune}. Comparing predictors within one search policy and region guiders within \promisetune{} keeps the surrounding selection logic fixed.

We compare \jev{} with conventional methods to determine whether these integrations improve on established tuning approaches. Since large language models (LLMs) can also select configurations and predict performance~\cite{llambo}, we include \deepseek{} (\deepseekmodel{})~\cite{deepseekv4} to test whether a decision model offers a quality--cost advantage over an LLM performing the same role. Both comparisons measure attained performance, decision time, and API fees across training and inference tasks, with held-out error as auxiliary evidence for prediction.

As an optimizer, \jev{} ranks below \flash{}, the strongest conventional optimizer across our tasks, in six of eight comparisons. Similarly, predictor-guided search with \jev{} ranks below the best conventional predictor on seven tasks. Region guidance offers a more promising integration, with \jev{} finding lower-perplexity configurations than the native tuner and both conventional guider variants in both GPU contexts, at additional decision time and API fees.

Building on these findings, our contributions are as follows.
\begin{enumerate}
\item We compare a general-purpose decision model, an LLM, and conventional methods across optimizer, performance-predictor, and region-guider roles under matched evaluation budgets.
\item We identify role-dependent quality--cost trade-offs and relate them to proposal repetition, candidate-ordering errors, and region preferences, deriving recommendations for model use in tuning.
\item We release measured trajectories, model outputs, and analysis code to reproduce the comparisons and evaluate further models.
\end{enumerate}

\section{Background}
\label{sec:background}
\subsection{Configuration Tuning}
Let $\mathcal{X}$ be a configuration space and $f(x)$ the measured performance of configuration $x$, with lower values preferred. After $t$ evaluations, $\mathcal{D}_t$ contains observed configuration--performance pairs and $\mathcal{U}_t=\mathcal{X}\setminus\{x_1,\ldots,x_t\}$ contains unseen configurations. Each evaluation updates the observations, and tuning reports the best performance within budget $B$,
\begin{align}
\mathcal{D}_{t+1}&=\mathcal{D}_t\cup\{(x_{t+1},f(x_{t+1}))\},
\label{eq:training-update}\\
y_B^{\mathrm{best}}&=\min_{1\leq i\leq B} f(x_i), \qquad x_i\in\mathcal{X}.
\label{eq:tuning-quality}
\end{align}
Three decision roles support this evaluation loop.

\subsection{Configuration Optimizer}
\label{sec:background-optimizer}
An optimizer supplies a policy $\pi$ that selects the next configuration from the history and unseen space,
\begin{equation}
x_{t+1}=\pi(\mathcal{D}_t,\mathcal{U}_t),\qquad x_{t+1}\in\mathcal{U}_t.
\label{eq:optimizer-policy}
\end{equation}
The policy can sample configurations, as in \randommethod{}~\cite{random}, or combine predictions and search, as in \flash{} and \smac{}~\cite{flash,smac}. Our optimizer comparison assigns the decision model this complete selection role alongside established tuners.

\subsection{Performance Predictor}
\label{sec:background-predictor}
For conventional predictors, a learning algorithm $\mathcal{A}$ fits a performance predictor $\hat f_t$ to observed pairs~\cite{gongdeep}. Its estimate $\hat f_t(x)$ guides selection from a candidate pool $\mathcal{C}_t\subseteq\mathcal{U}_t$. On prediction-guided steps, our shared policy uses
\begin{align}
\hat f_t&=\mathcal{A}(\mathcal{D}_t),
\label{eq:performance-learning}\\
x_{t+1}&\in\underset{x\in\mathcal{C}_t}{\arg\min}\;\hat f_t(x).
\label{eq:predictor-selection}
\end{align}
New observations update the conventional predictor's training set or the model's input history. Holding candidate sampling and exploration fixed lets us relate prediction error to tuning quality, whose method orderings can differ~\cite{accuracycanlie}.

\subsection{Region Guider}
\label{sec:background-region}
A region guider identifies promising subsets for a tuner~\cite{promisetune}. At a guidance update, $G$ maps observations to a region $\mathcal{R}\subseteq\mathcal{X}$, which the tuner's policy $\pi_T$ uses for selection,
\begin{equation}
\mathcal{R}=G(\mathcal{D}_t),\qquad x_{t+1}=\pi_T(\mathcal{D}_t,\mathcal{U}_t,\mathcal{R}).
\label{eq:region-guidance}
\end{equation}
Feature bounds describe subregions whose union forms $\mathcal{R}$. The tuner retains its predictor and acquisition rule, reusing the recommended region between guidance updates. Disabling guidance leaves the space unrestricted. Replacing $G$ within \promisetune{} lets us compare region advice while retaining the tuner's selection mechanism.

\subsection{General-Purpose Decision Models}
\label{sec:decision-models}
General-purpose decision models answer instruction-defined questions about an input state through typed outputs. \jev{}, TypeSafe AI's deployed System One model, supports option choices, rubric scores, and binary judgments~\cite{jevintro}, returning probabilities and confidence~\cite{jevconfidence}. These outputs let one model implement all three interfaces above. Extending classification and agent evaluations~\cite{jevbenchmark,jevworkflow}, we compare their effects on attained tuning performance and cost.

\section{Experiment Setup}
\label{sec:design}

We evaluate these three roles through the following research questions.
\begin{itemize}
\item \textbf{RQ1. How effective is a decision model as a configuration optimizer?} We compare its selected configurations and decision costs with conventional optimizers and \deepseek{} under matched budgets.
\item \textbf{RQ2. How effective is a decision model as a performance predictor for tuning?} We vary the predictor within a shared search policy and compare tuning outcomes, costs, and auxiliary held-out errors.
\item \textbf{RQ3. How effective is a decision model as a region guider?} We compare region-guidance alternatives within \promisetune{} through attained performance and decision costs.
\end{itemize}

\subsection{Systems, Configurations, and Performance}
\label{sec:data}
We select LCBench and HW-GPT-Bench for matched comparisons using recorded training and inference measurements~\cite{lcbench,hwgpt}. Pairing execution time with predictive quality tests whether tuning benefits both aspects of performance. LCBench supplies training time and validation error for MLPs trained from scratch for 50 epochs on \textsc{Adult} and \textsc{Covertype}. HW-GPT-Bench supplies inference latency and perplexity for small GPT-2-style architectures using inherited supernetwork weights on \textsc{A100} and \textsc{RTX2080} (Table~\ref{tab:data}). These four systems yield eight tuning tasks. Both GPUs share GPT configurations and perplexity measurements but have distinct latencies. During tuning, table lookup reveals only the selected configuration's measurement.

\begin{table*}[t]
\caption{Configuration spaces and performance measures of the four AI systems.}
\label{tab:data}
\centering\footnotesize\setlength{\tabcolsep}{3pt}
\renewcommand{\arraystretch}{1.15}
\begin{tabularx}{\textwidth}{@{}lcc>{\raggedright\arraybackslash}Xllr@{}}
\toprule
\textbf{System} & \textbf{\# Configs} & \textbf{\# Options} & \textbf{Option names} & \textbf{Performance} & \textbf{Description} & \textbf{Range} \\
\midrule
\rowcolor{black!7}
 & & & Batch size, learning rate, max. dropout, & Training time (s) & 50 epochs, from scratch & [71.7, 3\,684.0] \\
\hhline{>{\arrayrulecolor{black!7}}*{4}{-}>{\arrayrulecolor{black}}*{3}{-}}
\rowcolor{black!7}
\multirow{-2}{*}{\textsc{Adult}~\cite{lcbench}} & \multirow{-2}{*}{2,000} & \multirow{-2}{*}{7} & momentum, weight decay, max. units, layers & Val. error (\%) & Misclassification rate & [16.8, 75.7] \\
\hline
\multirow{2}{*}{\textsc{Covertype}~\cite{lcbench}} & \multirow{2}{*}{2,000} & \multirow{2}{*}{7} & Batch size, learning rate, max. dropout, & Training time (s) & 50 epochs, from scratch & [539.5, 45\,914.0] \\
\cline{5-7}
 & & & momentum, weight decay, max. units, layers & Val. error (\%) & Misclassification rate & [22.7, 94.8] \\
\hline
\rowcolor{black!7}
 & & & Depth, embedding, per-layer heads/MLP ratio; & Latency (ms) & \textsc{GPT-S} inference time & [9.4, 29.9] \\
\hhline{>{\arrayrulecolor{black!7}}*{4}{-}>{\arrayrulecolor{black}}*{3}{-}}
\rowcolor{black!7}
\multirow{-2}{*}{\textsc{A100}~\cite{hwgpt}} & \multirow{-2}{*}{10,000} & \multirow{-2}{*}{27} & bias/optional settings (one-hot encoded) & Perplexity & Next-token uncertainty & [22.2, 41.9] \\
\hline
\multirow{2}{*}{\textsc{RTX2080}~\cite{hwgpt}} & \multirow{2}{*}{10,000} & \multirow{2}{*}{27} & Depth, embedding, per-layer heads/MLP ratio; & Latency (ms) & \textsc{GPT-S} inference time & [18.2, 87.3] \\
\cline{5-7}
 & & & bias/optional settings (one-hot encoded) & Perplexity & Next-token uncertainty & [22.2, 41.9] \\
\bottomrule
\end{tabularx}
\end{table*}

Encoding affects performance learning~\cite{gongencoding}. We fix the feature encoding for each system and scale its numerical features to $[0,1]$. MLPs have seven numerical features. GPT has three global options (depth, embedding size, bias) and two per layer (attention heads, MLP expansion ratio). Its 10--12 layers give 23--27 active options. Layers 11 and 12 are optional; their settings include an absent category when depth is smaller. One-hot encoding these settings and bias yields 18 indicators alongside 22 numerical features. The resulting inputs have 7 dimensions for MLPs and 40 for GPT architectures.

\subsection{Models and Search Protocol}
\label{sec:roles}
\label{sec:protocol}
Following configuration-tuning studies~\cite{promisetune,accuracycanlie}, we budget distinct evaluated configurations to equalize measurement effort across methods. We use 100 evaluations, one of \promisetune{}'s studied budgets, with its ten-sample initialization and 30-run repetition count~\cite{promisetune}. This budget inspects 5\% of each MLP table and 1\% of each GPT table. Seeds 0--29 pair the initial samples across methods to control starting-point variation. Each run then makes 90 unseen choices, updating its observations through Eq.~\eqref{eq:training-update}. Checkpoints at 20, 40, 60, 80, and 100 evaluations expose gains from additional measurements.

We use \jevmodel{} and \deepseek{} with thinking disabled and temperature zero for the latter. Both receive task and parameter descriptions in a common format, together with their own complete observed history. The artifact provides complete request specifications for the 720 search runs per model.

\textbf{Execution hardware.} Conventional runs used 45 CPU nodes, requesting one CPU and 8 GiB per job with single-threaded OpenMP and BLAS. Their 30-seed timing intervals include node variation; the artifact supplies anonymous hardware records. API intervals include network and remote-inference latency.

\subsection{Quality, Cost, and Statistical Analysis}
\label{sec:analysis}
We prioritize best observed performance (Eq.~\eqref{eq:tuning-quality}) because prediction accuracy and tuning quality can yield different method orderings~\cite{accuracycanlie}. Held-out error supplies auxiliary evidence in RQ2. We report mean $\pm$ SD and Scott--Knott ESD ranks using unrounded per-seed values. This effect-aware grouping follows configuration-tuning practice~\cite{promisetune,gongdeep}; package 2.0.3 partitions means using Cohen's $d$~\cite{sk}. Rank 1 is best within each role--task comparison. API fees are ranked separately between the two models using their 30-seed vectors. For aggregate rank-1 counts, each GPU perplexity comparison has weight $1/2$ and each other comparison has weight 1, totaling seven. This gives the shared perplexity landscape one vote (Figure~\ref{fig:budget}). Paired comparisons use Wilcoxon tests, Holm adjustment, and Vargha--Delaney effect sizes; the artifact specifies each comparison family.

For RQ1, decision time measures the conventional selector or all API response intervals used to obtain each model decision. For RQ2 and RQ3, it measures the complete recorded selection process for every method, including tuner computation and API, accounting, pacing, and storage work within that interval. We sum decision time over the 90 selections per run and average across 30 seeds. Figure~\ref{fig:tradeoff} pairs the performance rank with mean time; lower values are preferred on both axes. Each front contains methods for which no competitor has both an equal or better rank and equal or lower time, with at least one strict improvement. Horizontal 95\% bootstrap intervals use 10,000 resamples of the run totals.

Conventional methods incur no API fees. For \jev{} and \deepseek{}, we sum all recorded API charges incurred by each run, converted to US cents at CNY 6.672332 per USD. Reused responses retain their original time and usage. The artifact records the pricing used in these calculations.

\section{Results and Analysis}
\label{sec:results}
\subsection{RQ1. Replacing the Optimizer}
\label{sec:rq1}
\subsubsection{Method}
We assess \jev{} as a complete optimizer against methods that accept configuration measurements and support single-metric search at a fixed evaluation fidelity. This matches our recorded endpoint data; multi-fidelity scheduling requires measurements at additional resource levels~\cite{lcbench}. The conventional baselines cover the sampling, tree-based, Bayesian, and causal strategies evaluated in configuration-tuning research~\cite{promisetune}.
\begin{itemize}
\item \randommethod{} samples unseen configurations uniformly. Its established role in hyperparameter optimization makes it the reference for gains from adaptive selection~\cite{random}.
\item \flash{} refits a CART tree and selects the lowest predicted value. We include this configuration-specific tuner for its measurement-efficient tree-guided search~\cite{flash}.
\item \smac{} provides random-forest Bayesian optimization through its hyperparameter-optimization interface (version 2.3.1). It represents an established optimizer for mixed configuration spaces~\cite{smac}.
\item \hebo{} combines a warped Gaussian process with ensemble acquisition (version 0.3.6). Its strong results on 108 hyperparameter tasks motivate this complementary Bayesian baseline~\cite{hebojournal}.
\item \promisetune{} adds causally filtered region rules to search~\cite{promisetune}. We use the authors' implementation, updating rules at each decision, to include an explicit region-guided optimizer.
\item \deepseek{} provides the LLM comparison under the same structured interface as \jev{}, isolating the model choice within that interface.
\end{itemize}

The models propose complete configurations through structured choices, with at most 21 ordered values per numeric parameter. We map proposals to the nearest unseen table configuration using normalized squared numeric distance, categorical mismatch distance, and seeded tie-breaking. For \smac{}, \hebo{}, and \promisetune{}, projection uses the scaled feature space. \smac{} and \hebo{} receive the mapped configuration's performance as feedback for the original proposal.

\begin{table*}[t]
\caption{(RQ1) Tuning quality of independent optimizers. $r$ denotes Scott--Knott ESD rank. Darker shades indicate better ranks within each comparison; ties share a shade and bold marks rank 1.}
\label{tab:optimizer}
\centering\footnotesize\setlength{\tabcolsep}{2.4pt}
\begingroup
\renewcommand{\arraystretch}{1.0}
\begin{tabular}{@{}llrcrcrcrcrcrcrc}
\toprule
\multirow{2}{*}{\textbf{System}} & \multirow{2}{*}{\textbf{Performance}} & \multicolumn{2}{c}{\textbf{\randommethod{}}} & \multicolumn{2}{c}{\textbf{\flash{}}} & \multicolumn{2}{c}{\textbf{\smac{}}} & \multicolumn{2}{c}{\textbf{\hebo{}}} & \multicolumn{2}{c}{\textbf{\promisetune{}}} & \multicolumn{2}{c}{\textbf{\deepseek{}}} & \multicolumn{2}{c}{\textbf{\jev{}}} \\
\cmidrule(lr){3-4}\cmidrule(lr){5-6}\cmidrule(lr){7-8}\cmidrule(lr){9-10}\cmidrule(lr){11-12}\cmidrule(lr){13-14}\cmidrule(lr){15-16}
 & & \textbf{Mean $\pm$ SD} & \textbf{$r$} & \textbf{Mean $\pm$ SD} & \textbf{$r$} & \textbf{Mean $\pm$ SD} & \textbf{$r$} & \textbf{Mean $\pm$ SD} & \textbf{$r$} & \textbf{Mean $\pm$ SD} & \textbf{$r$} & \textbf{Mean $\pm$ SD} & \textbf{$r$} & \textbf{Mean $\pm$ SD} & \textbf{$r$} \\
\midrule
\multirow{2}{*}{\textsc{Adult}} & Training time (s) & \metriccell{tabletraining}{5.0}{75.9\,{\scriptsize$\pm$\,2.9}} & \metriccell{tabletraining}{5.0}{5} & \metriccell{tabletraining}{45.0}{\textbf{72.0\,{\scriptsize$\pm$\,0.7}}} & \metriccell{tabletraining}{45.0}{\textbf{1}} & \metriccell{tabletraining}{15.0}{73.4\,{\scriptsize$\pm$\,1.8}} & \metriccell{tabletraining}{15.0}{4} & \metriccell{tabletraining}{25.0}{73.1\,{\scriptsize$\pm$\,1.6}} & \metriccell{tabletraining}{25.0}{3} & \metriccell{tabletraining}{25.0}{73.0\,{\scriptsize$\pm$\,1.2}} & \metriccell{tabletraining}{25.0}{3} & \metriccell{tabletraining}{35.0}{72.3\,{\scriptsize$\pm$\,0.2}} & \metriccell{tabletraining}{35.0}{2} & \metriccell{tabletraining}{15.0}{73.2\,{\scriptsize$\pm$\,1.7}} & \metriccell{tabletraining}{15.0}{4} \\
 & Val. error (\%) & \metriccell{tableerror}{31.7}{18.5\,{\scriptsize$\pm$\,1.0}} & \metriccell{tableerror}{31.7}{2} & \metriccell{tableerror}{45.0}{\textbf{18.1\,{\scriptsize$\pm$\,1.2}}} & \metriccell{tableerror}{45.0}{\textbf{1}} & \metriccell{tableerror}{18.3}{18.9\,{\scriptsize$\pm$\,0.8}} & \metriccell{tableerror}{18.3}{3} & \metriccell{tableerror}{5.0}{19.0\,{\scriptsize$\pm$\,0.7}} & \metriccell{tableerror}{5.0}{4} & \metriccell{tableerror}{18.3}{18.8\,{\scriptsize$\pm$\,0.7}} & \metriccell{tableerror}{18.3}{3} & \metriccell{tableerror}{5.0}{19.1\,{\scriptsize$\pm$\,0.6}} & \metriccell{tableerror}{5.0}{4} & \metriccell{tableerror}{18.3}{18.9\,{\scriptsize$\pm$\,0.8}} & \metriccell{tableerror}{18.3}{3} \\
\hline
\multirow{2}{*}{\textsc{Covertype}} & Training time (s) & \metriccell{tabletraining}{5.0}{597.5\,{\scriptsize$\pm$\,25.3}} & \metriccell{tabletraining}{5.0}{4} & \metriccell{tabletraining}{31.7}{557.1\,{\scriptsize$\pm$\,18.7}} & \metriccell{tabletraining}{31.7}{2} & \metriccell{tabletraining}{18.3}{567.3\,{\scriptsize$\pm$\,15.3}} & \metriccell{tabletraining}{18.3}{3} & \metriccell{tabletraining}{18.3}{569.0\,{\scriptsize$\pm$\,25.7}} & \metriccell{tabletraining}{18.3}{3} & \metriccell{tabletraining}{31.7}{558.0\,{\scriptsize$\pm$\,18.0}} & \metriccell{tabletraining}{31.7}{2} & \metriccell{tabletraining}{45.0}{\textbf{550.0\,{\scriptsize$\pm$\,3.0}}} & \metriccell{tabletraining}{45.0}{\textbf{1}} & \metriccell{tabletraining}{31.7}{559.4\,{\scriptsize$\pm$\,10.1}} & \metriccell{tabletraining}{31.7}{2} \\
 & Val. error (\%) & \metriccell{tableerror}{5.0}{29.7\,{\scriptsize$\pm$\,4.7}} & \metriccell{tableerror}{5.0}{5} & \metriccell{tableerror}{45.0}{\textbf{23.2\,{\scriptsize$\pm$\,1.8}}} & \metriccell{tableerror}{45.0}{\textbf{1}} & \metriccell{tableerror}{25.0}{26.1\,{\scriptsize$\pm$\,4.4}} & \metriccell{tableerror}{25.0}{3} & \metriccell{tableerror}{15.0}{27.2\,{\scriptsize$\pm$\,3.6}} & \metriccell{tableerror}{15.0}{4} & \metriccell{tableerror}{35.0}{24.8\,{\scriptsize$\pm$\,3.4}} & \metriccell{tableerror}{35.0}{2} & \metriccell{tableerror}{25.0}{25.5\,{\scriptsize$\pm$\,3.0}} & \metriccell{tableerror}{25.0}{3} & \metriccell{tableerror}{25.0}{25.7\,{\scriptsize$\pm$\,3.2}} & \metriccell{tableerror}{25.0}{3} \\
\hline
\multirow{2}{*}{\textsc{A100}} & Latency (ms) & \metriccell{tablelatency}{5.0}{10.5\,{\scriptsize$\pm$\,0.4}} & \metriccell{tablelatency}{5.0}{6} & \metriccell{tablelatency}{45.0}{\textbf{9.5\,{\scriptsize$\pm$\,0.1}}} & \metriccell{tablelatency}{45.0}{\textbf{1}} & \metriccell{tablelatency}{13.0}{10.3\,{\scriptsize$\pm$\,0.5}} & \metriccell{tablelatency}{13.0}{5} & \metriccell{tablelatency}{13.0}{10.3\,{\scriptsize$\pm$\,0.5}} & \metriccell{tablelatency}{13.0}{5} & \metriccell{tablelatency}{21.0}{10.0\,{\scriptsize$\pm$\,0.3}} & \metriccell{tablelatency}{21.0}{4} & \metriccell{tablelatency}{37.0}{9.6\,{\scriptsize$\pm$\,0.1}} & \metriccell{tablelatency}{37.0}{2} & \metriccell{tablelatency}{29.0}{9.7\,{\scriptsize$\pm$\,0.2}} & \metriccell{tablelatency}{29.0}{3} \\
 & Perplexity & \metriccell{tableperplexity}{13.0}{22.8\,{\scriptsize$\pm$\,0.3}} & \metriccell{tableperplexity}{13.0}{5} & \metriccell{tableperplexity}{45.0}{\textbf{22.3\,{\scriptsize$\pm$\,0.1}}} & \metriccell{tableperplexity}{45.0}{\textbf{1}} & \metriccell{tableperplexity}{21.0}{22.6\,{\scriptsize$\pm$\,0.2}} & \metriccell{tableperplexity}{21.0}{4} & \metriccell{tableperplexity}{29.0}{22.5\,{\scriptsize$\pm$\,0.2}} & \metriccell{tableperplexity}{29.0}{3} & \metriccell{tableperplexity}{37.0}{22.4\,{\scriptsize$\pm$\,0.2}} & \metriccell{tableperplexity}{37.0}{2} & \metriccell{tableperplexity}{45.0}{\textbf{22.3\,{\scriptsize$\pm$\,0.1}}} & \metriccell{tableperplexity}{45.0}{\textbf{1}} & \metriccell{tableperplexity}{5.0}{23.3\,{\scriptsize$\pm$\,0.4}} & \metriccell{tableperplexity}{5.0}{6} \\
\hline
\multirow{2}{*}{\textsc{RTX2080}} & Latency (ms) & \metriccell{tablelatency}{15.0}{21.2\,{\scriptsize$\pm$\,1.0}} & \metriccell{tablelatency}{15.0}{4} & \metriccell{tablelatency}{25.0}{19.4\,{\scriptsize$\pm$\,1.0}} & \metriccell{tablelatency}{25.0}{3} & \metriccell{tablelatency}{15.0}{21.4\,{\scriptsize$\pm$\,1.5}} & \metriccell{tablelatency}{15.0}{4} & \metriccell{tablelatency}{5.0}{21.7\,{\scriptsize$\pm$\,1.5}} & \metriccell{tablelatency}{5.0}{5} & \metriccell{tablelatency}{15.0}{21.2\,{\scriptsize$\pm$\,1.5}} & \metriccell{tablelatency}{15.0}{4} & \metriccell{tablelatency}{45.0}{\textbf{18.2\,{\scriptsize$\pm$\,0.0}}} & \metriccell{tablelatency}{45.0}{\textbf{1}} & \metriccell{tablelatency}{35.0}{19.0\,{\scriptsize$\pm$\,1.4}} & \metriccell{tablelatency}{35.0}{2} \\
 & Perplexity & \metriccell{tableperplexity}{13.0}{22.8\,{\scriptsize$\pm$\,0.3}} & \metriccell{tableperplexity}{13.0}{5} & \metriccell{tableperplexity}{45.0}{\textbf{22.3\,{\scriptsize$\pm$\,0.1}}} & \metriccell{tableperplexity}{45.0}{\textbf{1}} & \metriccell{tableperplexity}{21.0}{22.6\,{\scriptsize$\pm$\,0.2}} & \metriccell{tableperplexity}{21.0}{4} & \metriccell{tableperplexity}{29.0}{22.5\,{\scriptsize$\pm$\,0.2}} & \metriccell{tableperplexity}{29.0}{3} & \metriccell{tableperplexity}{37.0}{22.4\,{\scriptsize$\pm$\,0.2}} & \metriccell{tableperplexity}{37.0}{2} & \metriccell{tableperplexity}{45.0}{\textbf{22.3\,{\scriptsize$\pm$\,0.1}}} & \metriccell{tableperplexity}{45.0}{\textbf{1}} & \metriccell{tableperplexity}{5.0}{23.2\,{\scriptsize$\pm$\,0.4}} & \metriccell{tableperplexity}{5.0}{6} \\
\bottomrule
\end{tabular}
\endgroup

\end{table*}

\subsubsection{Results}
\textit{Tuning quality.} At 100 configurations, \flash{} attains rank 1 on six of eight tasks and \deepseek{} on four, while \jev{} attains none (Table~\ref{tab:optimizer}). \flash{} leads all validation-error and perplexity comparisons. \deepseek{} provides the strongest \textsc{RTX2080} latency result, reaching the recorded minimum of 18.2 ms in all 30 seeds, compared with 19.0 ms for \jev{}. The paired difference is $-0.84$ ms (95\% interval $[-1.34,-0.39]$, Holm-adjusted $p=0.0066$). These comparisons favor \flash{} for predictive quality and \deepseek{} for \textsc{RTX2080} latency.

\begin{figure*}[t]
\centering\includegraphics[width=\textwidth]{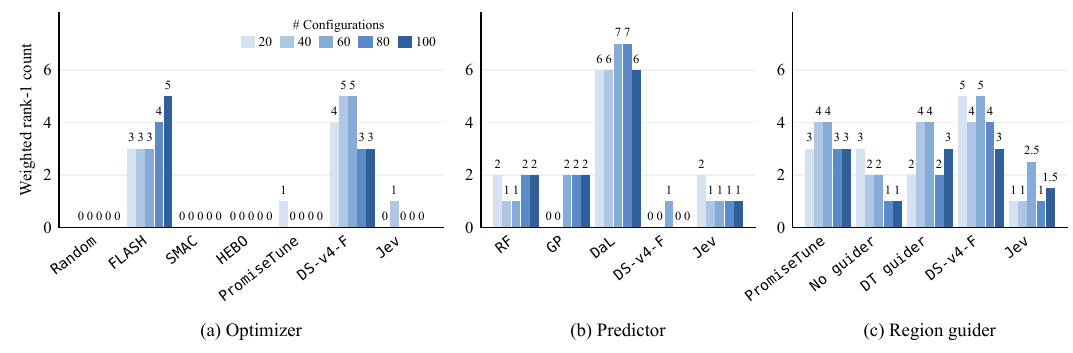}
\vspace{-0.9cm}
\caption{(RQ1--RQ3) Rank-1 counts across evaluation budgets for (a) optimizers, (b) predictors, and (c) region guiders. Bars use the shared-landscape weights defined in Section~\ref{sec:analysis}, including ties.}
\label{fig:budget}
\end{figure*}

\textit{Budget and cost.} Increasing the budget strengthens \flash{}'s position. From 20 to 100 configurations, its weighted rank-1 count grows from three to five, while \deepseek{}'s falls from four to three. \jev{} has one rank-1 result at 40 configurations and none at the other checkpoints (Figure~\ref{fig:budget}(a)). At 100 configurations, \flash{} also lies on the rank--time front in all eight comparisons (Figure~\ref{fig:tradeoff}).

Between the two API models, lower fees for \jev{} usually accompany weaker tuning quality. It costs 54--83\% less than \deepseek{} but has worse mean performance on seven tasks. \textsc{Adult} validation error is the exception, with lower mean error and lower fees for \jev{} (Tables~\ref{tab:optimizer} and~\ref{tab:region}).

\begin{rqanswer}
\textbf{RQ1 Answer.} For independent optimization, \flash{} is the stronger default, reaching rank~1 on six of eight tasks. \jev{} offers lower API fees than \deepseek{}, usually at the expense of tuning quality.
\end{rqanswer}

\subsection{RQ2. Replacing the Performance Predictor}
\label{sec:rq2}
\subsubsection{Method}
We next assess \jev{} as a performance predictor while fixing candidate generation and selection. Prior evidence shows that the predictor and search policy jointly affect tuning quality~\cite{accuracycanlie}; this shared loop isolates the predictor choice. We select complementary global and local models used in configuration-performance learning~\cite{gongdeep}.
\begin{itemize}
\item \rf{} averages 100 regression trees~\cite{rf}. It represents the tree-ensemble predictors used by configuration tuners such as \smac{}~\cite{smac}.
\item \gp{} uses a Mat\'ern-5/2 covariance, white noise, and normalized labels~\cite{gp}. It supplies a smooth probabilistic alternative to tree-based prediction.
\item \dal{} adaptively partitions observations and fits local hierarchical neural predictors. We select the authors' extended implementation because it targets sparse configuration landscapes and has been evaluated across 12 systems~\cite{daljournal}.
\item \deepseek{} receives the same candidate and scoring interface as \jev{}, enabling a matched comparison of the two models.
\end{itemize}

At each step, we sample 64 unseen candidates to bound prediction work equally across methods. The loop selects the lowest prediction on nine of every ten steps and explores on the remaining step through a shared seeded schedule. On prediction-guided steps, conventional predictors refit using all observations, and the two models receive the updated history. \jev{} uses Score questions with five equally spaced levels across the observed range. We map its fractional score $s\in[0,4]$ to $\hat y=y_{\min}+s(y_{\max}-y_{\min})/4$, using the same scale for \deepseek{}. Table~\ref{tab:predictor} (left) reports search quality; Figures~\ref{fig:budget}(b) and~\ref{fig:tradeoff} relate it to budget and decision time.

An auxiliary test compares numerical accuracy using MAPE, a scale-relative measure widely used in configuration-performance learning~\cite{accuracycanlie}. Each seed supplies all five predictors with the same 100 training configurations and 128 disjoint test configurations, holding sample size and test difficulty fixed. Table~\ref{tab:predictor} (right) reports $100\times128^{-1}\sum_{i=1}^{128}|\hat y_i-y_i|/|y_i|$ on positive labels. To assess the effect of the models' bounded output range, we also clip conventional predictions to the training-label range and recompute errors. The artifact includes complementary MAE and RMSE.

\begin{table*}[t]
\caption{(RQ2) Predictor-guided tuning quality (left) and auxiliary held-out MAPE (\%) (right). Formatting follows Table~\ref{tab:optimizer}.}
\label{tab:predictor}
\centering\scriptsize\setlength{\tabcolsep}{1.0pt}
\resizebox{\textwidth}{!}{\begingroup
\renewcommand{\arraystretch}{1.0}
\begin{tabular}{@{}llrcrcrcrcrc||rcrcrcrcrc}
\toprule
\multirow{2}{*}{\textbf{System}} & \multirow{2}{*}{\textbf{Performance}} & \multicolumn{2}{c}{\textbf{\rf{}}} & \multicolumn{2}{c}{\textbf{\gp{}}} & \multicolumn{2}{c}{\textbf{\dal{}}} & \multicolumn{2}{c}{\textbf{\deepseek{}}} & \multicolumn{2}{c||}{\textbf{\jev{}}} & \multicolumn{2}{c}{\textbf{\rf{}}} & \multicolumn{2}{c}{\textbf{\gp{}}} & \multicolumn{2}{c}{\textbf{\dal{}}} & \multicolumn{2}{c}{\textbf{\deepseek{}}} & \multicolumn{2}{c}{\textbf{\jev{}}} \\
\cmidrule(lr){3-4}\cmidrule(lr){5-6}\cmidrule(lr){7-8}\cmidrule(lr){9-10}\cmidrule(lr){11-12}\cmidrule(lr){13-14}\cmidrule(lr){15-16}\cmidrule(lr){17-18}\cmidrule(lr){19-20}\cmidrule(lr){21-22}
 & & \textbf{Mean$\pm$SD} & \textbf{$r$} & \textbf{Mean$\pm$SD} & \textbf{$r$} & \textbf{Mean$\pm$SD} & \textbf{$r$} & \textbf{Mean$\pm$SD} & \textbf{$r$} & \textbf{Mean$\pm$SD} & \textbf{$r$} & \textbf{Mean$\pm$SD} & \textbf{$r$} & \textbf{Mean$\pm$SD} & \textbf{$r$} & \textbf{Mean$\pm$SD} & \textbf{$r$} & \textbf{Mean$\pm$SD} & \textbf{$r$} & \textbf{Mean$\pm$SD} & \textbf{$r$} \\
\midrule
\multirow{2}{*}{\textsc{Adult}} & Training time (s) & \metriccell{tabletraining}{31.7}{72.5\,{\scriptsize$\pm$\,1.2}} & \metriccell{tabletraining}{31.7}{2} & \metriccell{tabletraining}{31.7}{72.5\,{\scriptsize$\pm$\,1.5}} & \metriccell{tabletraining}{31.7}{2} & \metriccell{tabletraining}{45.0}{\textbf{72.0\,{\scriptsize$\pm$\,0.6}}} & \metriccell{tabletraining}{45.0}{\textbf{1}} & \metriccell{tabletraining}{5.0}{75.6\,{\scriptsize$\pm$\,2.7}} & \metriccell{tabletraining}{5.0}{4} & \metriccell{tabletraining}{18.3}{74.2\,{\scriptsize$\pm$\,2.3}} & \metriccell{tabletraining}{18.3}{3} & \metriccell{tabletraining}{31.7}{29.6\,{\scriptsize$\pm$\,22.6}} & \metriccell{tabletraining}{31.7}{2} & \metriccell{tabletraining}{31.7}{31.1\,{\scriptsize$\pm$\,20.4}} & \metriccell{tabletraining}{31.7}{2} & \metriccell{tabletraining}{45.0}{\textbf{20.0\,{\scriptsize$\pm$\,22.2}}} & \metriccell{tabletraining}{45.0}{\textbf{1}} & \metriccell{tabletraining}{18.3}{33.9\,{\scriptsize$\pm$\,4.0}} & \metriccell{tabletraining}{18.3}{3} & \metriccell{tabletraining}{5.0}{127.3\,{\scriptsize$\pm$\,70.0}} & \metriccell{tabletraining}{5.0}{4} \\
 & Val. error (\%) & \metriccell{tableerror}{45.0}{\textbf{18.7\,{\scriptsize$\pm$\,0.9}}} & \metriccell{tableerror}{45.0}{\textbf{1}} & \metriccell{tableerror}{45.0}{\textbf{18.8\,{\scriptsize$\pm$\,0.8}}} & \metriccell{tableerror}{45.0}{\textbf{1}} & \metriccell{tableerror}{5.0}{18.8\,{\scriptsize$\pm$\,0.7}} & \metriccell{tableerror}{5.0}{2} & \metriccell{tableerror}{5.0}{18.8\,{\scriptsize$\pm$\,0.8}} & \metriccell{tableerror}{5.0}{2} & \metriccell{tableerror}{5.0}{18.9\,{\scriptsize$\pm$\,0.9}} & \metriccell{tableerror}{5.0}{2} & \metriccell{tableerror}{45.0}{\textbf{7.6\,{\scriptsize$\pm$\,1.9}}} & \metriccell{tableerror}{45.0}{\textbf{1}} & \metriccell{tableerror}{25.0}{9.2\,{\scriptsize$\pm$\,1.3}} & \metriccell{tableerror}{25.0}{2} & \metriccell{tableerror}{25.0}{10.1\,{\scriptsize$\pm$\,8.3}} & \metriccell{tableerror}{25.0}{2} & \metriccell{tableerror}{5.0}{16.3\,{\scriptsize$\pm$\,4.3}} & \metriccell{tableerror}{5.0}{3} & \metriccell{tableerror}{5.0}{15.4\,{\scriptsize$\pm$\,4.6}} & \metriccell{tableerror}{5.0}{3} \\
\hline
\multirow{2}{*}{\textsc{Covertype}} & Training time (s) & \metriccell{tabletraining}{45.0}{\textbf{557.2\,{\scriptsize$\pm$\,13.1}}} & \metriccell{tabletraining}{45.0}{\textbf{1}} & \metriccell{tabletraining}{31.7}{561.4\,{\scriptsize$\pm$\,21.7}} & \metriccell{tabletraining}{31.7}{2} & \metriccell{tabletraining}{45.0}{\textbf{558.2\,{\scriptsize$\pm$\,15.3}}} & \metriccell{tabletraining}{45.0}{\textbf{1}} & \metriccell{tabletraining}{5.0}{593.2\,{\scriptsize$\pm$\,28.6}} & \metriccell{tabletraining}{5.0}{4} & \metriccell{tabletraining}{18.3}{569.8\,{\scriptsize$\pm$\,23.9}} & \metriccell{tabletraining}{18.3}{3} & \metriccell{tabletraining}{31.7}{45.6\,{\scriptsize$\pm$\,26.0}} & \metriccell{tabletraining}{31.7}{2} & \metriccell{tabletraining}{18.3}{51.9\,{\scriptsize$\pm$\,20.9}} & \metriccell{tabletraining}{18.3}{3} & \metriccell{tabletraining}{18.3}{50.3\,{\scriptsize$\pm$\,104.8}} & \metriccell{tabletraining}{18.3}{3} & \metriccell{tabletraining}{45.0}{\textbf{40.2\,{\scriptsize$\pm$\,15.7}}} & \metriccell{tabletraining}{45.0}{\textbf{1}} & \metriccell{tabletraining}{5.0}{141.1\,{\scriptsize$\pm$\,72.6}} & \metriccell{tabletraining}{5.0}{4} \\
 & Val. error (\%) & \metriccell{tableerror}{31.7}{23.4\,{\scriptsize$\pm$\,1.8}} & \metriccell{tableerror}{31.7}{2} & \metriccell{tableerror}{31.7}{23.8\,{\scriptsize$\pm$\,2.2}} & \metriccell{tableerror}{31.7}{2} & \metriccell{tableerror}{45.0}{\textbf{22.8\,{\scriptsize$\pm$\,0.2}}} & \metriccell{tableerror}{45.0}{\textbf{1}} & \metriccell{tableerror}{5.0}{28.6\,{\scriptsize$\pm$\,3.9}} & \metriccell{tableerror}{5.0}{4} & \metriccell{tableerror}{18.3}{27.2\,{\scriptsize$\pm$\,4.4}} & \metriccell{tableerror}{18.3}{3} & \metriccell{tableerror}{45.0}{\textbf{7.0\,{\scriptsize$\pm$\,0.8}}} & \metriccell{tableerror}{45.0}{\textbf{1}} & \metriccell{tableerror}{25.0}{11.0\,{\scriptsize$\pm$\,2.4}} & \metriccell{tableerror}{25.0}{2} & \metriccell{tableerror}{25.0}{12.7\,{\scriptsize$\pm$\,26.4}} & \metriccell{tableerror}{25.0}{2} & \metriccell{tableerror}{5.0}{24.5\,{\scriptsize$\pm$\,9.6}} & \metriccell{tableerror}{5.0}{3} & \metriccell{tableerror}{25.0}{12.2\,{\scriptsize$\pm$\,2.1}} & \metriccell{tableerror}{25.0}{2} \\
\hline
\multirow{2}{*}{\textsc{A100}} & Latency (ms) & \metriccell{tablelatency}{25.0}{9.7\,{\scriptsize$\pm$\,0.3}} & \metriccell{tablelatency}{25.0}{2} & \metriccell{tablelatency}{45.0}{\textbf{9.6\,{\scriptsize$\pm$\,0.2}}} & \metriccell{tablelatency}{45.0}{\textbf{1}} & \metriccell{tablelatency}{45.0}{\textbf{9.6\,{\scriptsize$\pm$\,0.2}}} & \metriccell{tablelatency}{45.0}{\textbf{1}} & \metriccell{tablelatency}{5.0}{10.2\,{\scriptsize$\pm$\,0.4}} & \metriccell{tablelatency}{5.0}{3} & \metriccell{tablelatency}{25.0}{9.7\,{\scriptsize$\pm$\,0.4}} & \metriccell{tablelatency}{25.0}{2} & \metriccell{tablelatency}{25.0}{3.6\,{\scriptsize$\pm$\,0.3}} & \metriccell{tablelatency}{25.0}{3} & \metriccell{tablelatency}{35.0}{3.5\,{\scriptsize$\pm$\,0.3}} & \metriccell{tablelatency}{35.0}{2} & \metriccell{tablelatency}{45.0}{\textbf{1.5\,{\scriptsize$\pm$\,0.3}}} & \metriccell{tablelatency}{45.0}{\textbf{1}} & \metriccell{tablelatency}{5.0}{14.7\,{\scriptsize$\pm$\,5.5}} & \metriccell{tablelatency}{5.0}{5} & \metriccell{tablelatency}{15.0}{8.6\,{\scriptsize$\pm$\,1.6}} & \metriccell{tablelatency}{15.0}{4} \\
 & Perplexity & \metriccell{tableperplexity}{31.7}{22.4\,{\scriptsize$\pm$\,0.1}} & \metriccell{tableperplexity}{31.7}{2} & \metriccell{tableperplexity}{31.7}{22.4\,{\scriptsize$\pm$\,0.2}} & \metriccell{tableperplexity}{31.7}{2} & \metriccell{tableperplexity}{45.0}{\textbf{22.3\,{\scriptsize$\pm$\,0.1}}} & \metriccell{tableperplexity}{45.0}{\textbf{1}} & \metriccell{tableperplexity}{18.3}{22.6\,{\scriptsize$\pm$\,0.2}} & \metriccell{tableperplexity}{18.3}{3} & \metriccell{tableperplexity}{5.0}{22.8\,{\scriptsize$\pm$\,0.3}} & \metriccell{tableperplexity}{5.0}{4} & \metriccell{tableperplexity}{35.0}{1.1\,{\scriptsize$\pm$\,0.1}} & \metriccell{tableperplexity}{35.0}{2} & \metriccell{tableperplexity}{25.0}{4.9\,{\scriptsize$\pm$\,0.4}} & \metriccell{tableperplexity}{25.0}{3} & \metriccell{tableperplexity}{45.0}{\textbf{0.7\,{\scriptsize$\pm$\,0.2}}} & \metriccell{tableperplexity}{45.0}{\textbf{1}} & \metriccell{tableperplexity}{5.0}{8.6\,{\scriptsize$\pm$\,4.1}} & \metriccell{tableperplexity}{5.0}{5} & \metriccell{tableperplexity}{15.0}{6.0\,{\scriptsize$\pm$\,1.1}} & \metriccell{tableperplexity}{15.0}{4} \\
\hline
\multirow{2}{*}{\textsc{RTX2080}} & Latency (ms) & \metriccell{tablelatency}{25.0}{19.6\,{\scriptsize$\pm$\,0.7}} & \metriccell{tablelatency}{25.0}{2} & \metriccell{tablelatency}{25.0}{19.5\,{\scriptsize$\pm$\,0.9}} & \metriccell{tablelatency}{25.0}{2} & \metriccell{tablelatency}{45.0}{\textbf{19.2\,{\scriptsize$\pm$\,1.0}}} & \metriccell{tablelatency}{45.0}{\textbf{1}} & \metriccell{tablelatency}{5.0}{20.7\,{\scriptsize$\pm$\,0.9}} & \metriccell{tablelatency}{5.0}{3} & \metriccell{tablelatency}{45.0}{\textbf{19.2\,{\scriptsize$\pm$\,0.8}}} & \metriccell{tablelatency}{45.0}{\textbf{1}} & \metriccell{tablelatency}{45.0}{\textbf{7.0\,{\scriptsize$\pm$\,0.6}}} & \metriccell{tablelatency}{45.0}{\textbf{1}} & \metriccell{tablelatency}{45.0}{\textbf{7.0\,{\scriptsize$\pm$\,0.7}}} & \metriccell{tablelatency}{45.0}{\textbf{1}} & \metriccell{tablelatency}{31.7}{7.8\,{\scriptsize$\pm$\,1.0}} & \metriccell{tablelatency}{31.7}{2} & \metriccell{tablelatency}{5.0}{19.8\,{\scriptsize$\pm$\,6.6}} & \metriccell{tablelatency}{5.0}{4} & \metriccell{tablelatency}{18.3}{12.0\,{\scriptsize$\pm$\,1.9}} & \metriccell{tablelatency}{18.3}{3} \\
 & Perplexity & \metriccell{tableperplexity}{31.7}{22.4\,{\scriptsize$\pm$\,0.1}} & \metriccell{tableperplexity}{31.7}{2} & \metriccell{tableperplexity}{31.7}{22.4\,{\scriptsize$\pm$\,0.2}} & \metriccell{tableperplexity}{31.7}{2} & \metriccell{tableperplexity}{45.0}{\textbf{22.3\,{\scriptsize$\pm$\,0.1}}} & \metriccell{tableperplexity}{45.0}{\textbf{1}} & \metriccell{tableperplexity}{18.3}{22.6\,{\scriptsize$\pm$\,0.2}} & \metriccell{tableperplexity}{18.3}{3} & \metriccell{tableperplexity}{5.0}{22.7\,{\scriptsize$\pm$\,0.3}} & \metriccell{tableperplexity}{5.0}{4} & \metriccell{tableperplexity}{35.0}{1.1\,{\scriptsize$\pm$\,0.1}} & \metriccell{tableperplexity}{35.0}{2} & \metriccell{tableperplexity}{25.0}{4.9\,{\scriptsize$\pm$\,0.4}} & \metriccell{tableperplexity}{25.0}{3} & \metriccell{tableperplexity}{45.0}{\textbf{0.7\,{\scriptsize$\pm$\,0.2}}} & \metriccell{tableperplexity}{45.0}{\textbf{1}} & \metriccell{tableperplexity}{5.0}{9.0\,{\scriptsize$\pm$\,4.0}} & \metriccell{tableperplexity}{5.0}{5} & \metriccell{tableperplexity}{15.0}{6.1\,{\scriptsize$\pm$\,1.2}} & \metriccell{tableperplexity}{15.0}{4} \\
\bottomrule
\end{tabular}
\endgroup
}
\end{table*}

\subsubsection{Results}
\textit{Predictor-guided tuning.} Conventional predictors rank above \jev{} on seven of eight tasks. On \textsc{RTX2080} latency, \jev{} and \dal{} share rank 1 at 19.2 ms (Table~\ref{tab:predictor}). This conventional advantage persists across budgets, with weighted rank-1 counts of six or seven for \dal{} at every checkpoint, compared with one or two for \jev{}. Conventional predictors occupy the rank--time fronts across all eight comparisons (Figure~\ref{fig:tradeoff}).

Between the two API models, \jev{} outranks \deepseek{} on all four efficiency tasks. Its \textsc{RTX2080} latency is 1.52 ms lower (95\% interval $[1.03,1.99]$; adjusted $p\leq0.0022$). This reverses their ordering in independent optimization, showing that their relative performance depends on the tuning role. \jev{} also incurs lower API fees on both GPU latency tasks (Table~\ref{tab:region}).

\textit{Prediction accuracy.} The auxiliary prediction test also favors conventional predictors. \jev{}'s MAPE is 1.7--9.0 times that of the best conventional predictor across the eight tasks; for perplexity, the values are 6.0--6.1\% for \jev{} and 0.7\% for \dal{}. Clipping conventional predictions to the same observed range preserves their advantage. However, the rank-1 \textsc{RTX2080} latency result for \jev{} shows that higher numerical error can accompany competitive tuning quality (Table~\ref{tab:predictor}). Candidate selection and held-out numerical error therefore provide complementary evidence about a predictor's usefulness.

\begin{rqanswer}
\textbf{RQ2 Answer.} Conventional predictors are the stronger choice for guiding search, outperforming \jev{} on seven of eight tasks. For GPU latency tuning, \jev{} offers better performance and lower API fees than \deepseek{}.
\end{rqanswer}

\subsection{RQ3. Replacing the Region Guider}
\label{sec:rq3}
\subsubsection{Method}
We assess \jev{} within \promisetune{} because its explicit region rules provide a replaceable guidance interface~\cite{promisetune}. Each variant retains the tuner's surrogate, acquisition calculation, candidate generator, and configuration-selection path. The guider steers candidate generation through feature bounds; the tuner selects a proposal and maps it to an unseen recorded configuration. We compare it with the original tuner, two component controls, and the LLM guider.
\begin{itemize}
\item The original \promisetune{} derives random-forest path rules and filters them through causal discovery and effect estimation~\cite{promisetune}. Reusing its RQ1 trajectories and timings establishes the native-tuner reference, with rules refreshed at every decision.
\item \noregion{} disables region rules to measure the tuner's performance when all candidate regions remain available. This ablation isolates the contribution of restricting search.
\item \treeregion{} recommends up to three lowest-mean leaves from a depth-three decision tree, with at least three observations per leaf. Inspired by tree-path region representations~\cite{promisetune}, this study control tests guidance learned directly from measured configurations.
\item \deepseek{} uses \jev{}'s region interface to compare model-generated guidance under identical output constraints.
\end{itemize}

Both models choose among eight predefined interval options for each encoded feature in each of three regions. The \treeregion{} and both models refresh every ten evaluations, giving them equal opportunities to react to new observations. This schedule yields nine API calls per run, reducing decision frequency relative to RQ1 and RQ2.

\begin{table*}[t]
\caption{(RQ3) Tuning quality of region guiders (left) and mean API fees in US cents for all three roles (right). Formatting follows Table~\ref{tab:optimizer}.}
\label{tab:region}
\centering\scriptsize\setlength{\tabcolsep}{1.8pt}
\resizebox{\textwidth}{!}{\begingroup
\renewcommand{\arraystretch}{1.0}
\begin{tabular}{@{}llrcrcrcrcrc||rcrc|rcrc|rcrc}
\toprule
\multirow{3}{*}{\textbf{System}} & \multirow{3}{*}{\textbf{Performance}} & \multicolumn{2}{c}{\multirow{2}{*}{\textbf{\promisetune{}}}} & \multicolumn{2}{c}{\multirow{2}{*}{\textbf{\noregion{}}}} & \multicolumn{2}{c}{\multirow{2}{*}{\textbf{\treeregion{}}}} & \multicolumn{2}{c}{\multirow{2}{*}{\textbf{\deepseek{}}}} & \multicolumn{2}{c||}{\multirow{2}{*}{\textbf{\jev{}}}} & \multicolumn{4}{c|}{\textbf{Optimizer API fee}} & \multicolumn{4}{c|}{\textbf{Predictor API fee}} & \multicolumn{4}{c}{\textbf{Region API fee}} \\
\cmidrule(lr){13-16}\cmidrule(lr){17-20}\cmidrule(lr){21-24}
 &  &  &  &  &  &  &  &  &  &  & \multicolumn{1}{c||}{} & \multicolumn{2}{c}{\textbf{\deepseek{}}} & \multicolumn{2}{c|}{\textbf{\jev{}}} & \multicolumn{2}{c}{\textbf{\deepseek{}}} & \multicolumn{2}{c|}{\textbf{\jev{}}} & \multicolumn{2}{c}{\textbf{\deepseek{}}} & \multicolumn{2}{c}{\textbf{\jev{}}} \\
\cmidrule(lr){3-4}\cmidrule(lr){5-6}\cmidrule(lr){7-8}\cmidrule(lr){9-10}\cmidrule(lr){11-12}\cmidrule(lr){13-14}\cmidrule(lr){15-16}\cmidrule(lr){17-18}\cmidrule(lr){19-20}\cmidrule(lr){21-22}\cmidrule(lr){23-24}
 & & \textbf{Mean $\pm$ SD} & \textbf{$r$} & \textbf{Mean $\pm$ SD} & \textbf{$r$} & \textbf{Mean $\pm$ SD} & \textbf{$r$} & \textbf{Mean $\pm$ SD} & \textbf{$r$} & \textbf{Mean $\pm$ SD} & \textbf{$r$} & \textbf{Mean} & \textbf{$r$} & \textbf{Mean} & \textbf{$r$} & \textbf{Mean} & \textbf{$r$} & \textbf{Mean} & \textbf{$r$} & \textbf{Mean} & \textbf{$r$} & \textbf{Mean} & \textbf{$r$} \\
\midrule
\multirow{2}{*}{\textsc{Adult}} & Training time (s) & \metriccell{tabletraining}{45.0}{\textbf{73.0\,{\scriptsize$\pm$\,1.2}}} & \metriccell{tabletraining}{45.0}{\textbf{1}} & \metriccell{tabletraining}{5.0}{74.1\,{\scriptsize$\pm$\,1.8}} & \metriccell{tabletraining}{5.0}{3} & \metriccell{tabletraining}{25.0}{73.6\,{\scriptsize$\pm$\,1.9}} & \metriccell{tabletraining}{25.0}{2} & \metriccell{tabletraining}{25.0}{73.5\,{\scriptsize$\pm$\,1.4}} & \metriccell{tabletraining}{25.0}{2} & \metriccell{tabletraining}{5.0}{73.9\,{\scriptsize$\pm$\,2.8}} & \metriccell{tabletraining}{5.0}{3} & \metriccell{tablecost}{5.0}{9.0} & \metriccell{tablecost}{5.0}{2} & \metriccell{tablecost}{45.0}{\textbf{4.2}} & \metriccell{tablecost}{45.0}{\textbf{1}} & \metriccell{tablecost}{45.0}{\textbf{9.8}} & \metriccell{tablecost}{45.0}{\textbf{1}} & \metriccell{tablecost}{5.0}{10.0} & \metriccell{tablecost}{5.0}{2} & \metriccell{tablecost}{5.0}{1.9} & \metriccell{tablecost}{5.0}{2} & \metriccell{tablecost}{45.0}{\textbf{0.7}} & \metriccell{tablecost}{45.0}{\textbf{1}} \\
 & Val. error (\%) & \metriccell{tableerror}{45.0}{\textbf{18.8\,{\scriptsize$\pm$\,0.7}}} & \metriccell{tableerror}{45.0}{\textbf{1}} & \metriccell{tableerror}{45.0}{\textbf{18.8\,{\scriptsize$\pm$\,0.8}}} & \metriccell{tableerror}{45.0}{\textbf{1}} & \metriccell{tableerror}{45.0}{\textbf{18.8\,{\scriptsize$\pm$\,0.9}}} & \metriccell{tableerror}{45.0}{\textbf{1}} & \metriccell{tableerror}{5.0}{19.2\,{\scriptsize$\pm$\,0.6}} & \metriccell{tableerror}{5.0}{2} & \metriccell{tableerror}{45.0}{\textbf{18.7\,{\scriptsize$\pm$\,1.2}}} & \metriccell{tableerror}{45.0}{\textbf{1}} & \metriccell{tablecost}{5.0}{9.1} & \metriccell{tablecost}{5.0}{2} & \metriccell{tablecost}{45.0}{\textbf{4.2}} & \metriccell{tablecost}{45.0}{\textbf{1}} & \metriccell{tablecost}{45.0}{\textbf{9.8}} & \metriccell{tablecost}{45.0}{\textbf{1}} & \metriccell{tablecost}{5.0}{10.1} & \metriccell{tablecost}{5.0}{2} & \metriccell{tablecost}{5.0}{1.9} & \metriccell{tablecost}{5.0}{2} & \metriccell{tablecost}{45.0}{\textbf{0.7}} & \metriccell{tablecost}{45.0}{\textbf{1}} \\
\hline
\multirow{2}{*}{\textsc{Covertype}} & Training time (s) & \metriccell{tabletraining}{25.0}{558.0\,{\scriptsize$\pm$\,18.0}} & \metriccell{tabletraining}{25.0}{2} & \metriccell{tabletraining}{25.0}{559.9\,{\scriptsize$\pm$\,21.3}} & \metriccell{tabletraining}{25.0}{2} & \metriccell{tabletraining}{45.0}{\textbf{555.3\,{\scriptsize$\pm$\,16.4}}} & \metriccell{tabletraining}{45.0}{\textbf{1}} & \metriccell{tabletraining}{25.0}{557.6\,{\scriptsize$\pm$\,18.3}} & \metriccell{tabletraining}{25.0}{2} & \metriccell{tabletraining}{5.0}{572.5\,{\scriptsize$\pm$\,30.9}} & \metriccell{tabletraining}{5.0}{3} & \metriccell{tablecost}{5.0}{9.0} & \metriccell{tablecost}{5.0}{2} & \metriccell{tablecost}{45.0}{\textbf{4.2}} & \metriccell{tablecost}{45.0}{\textbf{1}} & \metriccell{tablecost}{45.0}{\textbf{9.8}} & \metriccell{tablecost}{45.0}{\textbf{1}} & \metriccell{tablecost}{5.0}{10.0} & \metriccell{tablecost}{5.0}{2} & \metriccell{tablecost}{5.0}{1.9} & \metriccell{tablecost}{5.0}{2} & \metriccell{tablecost}{45.0}{\textbf{0.7}} & \metriccell{tablecost}{45.0}{\textbf{1}} \\
 & Val. error (\%) & \metriccell{tableerror}{45.0}{\textbf{24.8\,{\scriptsize$\pm$\,3.4}}} & \metriccell{tableerror}{45.0}{\textbf{1}} & \metriccell{tableerror}{25.0}{25.6\,{\scriptsize$\pm$\,3.0}} & \metriccell{tableerror}{25.0}{2} & \metriccell{tableerror}{45.0}{\textbf{24.6\,{\scriptsize$\pm$\,2.5}}} & \metriccell{tableerror}{45.0}{\textbf{1}} & \metriccell{tableerror}{5.0}{27.5\,{\scriptsize$\pm$\,4.5}} & \metriccell{tableerror}{5.0}{3} & \metriccell{tableerror}{25.0}{25.5\,{\scriptsize$\pm$\,3.3}} & \metriccell{tableerror}{25.0}{2} & \metriccell{tablecost}{5.0}{9.1} & \metriccell{tablecost}{5.0}{2} & \metriccell{tablecost}{45.0}{\textbf{4.2}} & \metriccell{tablecost}{45.0}{\textbf{1}} & \metriccell{tablecost}{45.0}{\textbf{9.9}} & \metriccell{tablecost}{45.0}{\textbf{1}} & \metriccell{tablecost}{5.0}{10.1} & \metriccell{tablecost}{5.0}{2} & \metriccell{tablecost}{5.0}{1.9} & \metriccell{tablecost}{5.0}{2} & \metriccell{tablecost}{45.0}{\textbf{0.7}} & \metriccell{tablecost}{45.0}{\textbf{1}} \\
\hline
\multirow{2}{*}{\textsc{A100}} & Latency (ms) & \metriccell{tablelatency}{25.0}{10.0\,{\scriptsize$\pm$\,0.3}} & \metriccell{tablelatency}{25.0}{2} & \metriccell{tablelatency}{5.0}{10.2\,{\scriptsize$\pm$\,0.3}} & \metriccell{tablelatency}{5.0}{3} & \metriccell{tablelatency}{25.0}{10.0\,{\scriptsize$\pm$\,0.4}} & \metriccell{tablelatency}{25.0}{2} & \metriccell{tablelatency}{45.0}{\textbf{9.8\,{\scriptsize$\pm$\,0.4}}} & \metriccell{tablelatency}{45.0}{\textbf{1}} & \metriccell{tablelatency}{5.0}{10.1\,{\scriptsize$\pm$\,0.3}} & \metriccell{tablelatency}{5.0}{3} & \metriccell{tablecost}{5.0}{19.1} & \metriccell{tablecost}{5.0}{2} & \metriccell{tablecost}{45.0}{\textbf{3.3}} & \metriccell{tablecost}{45.0}{\textbf{1}} & \metriccell{tablecost}{5.0}{13.4} & \metriccell{tablecost}{5.0}{2} & \metriccell{tablecost}{45.0}{\textbf{9.0}} & \metriccell{tablecost}{45.0}{\textbf{1}} & \metriccell{tablecost}{5.0}{5.9} & \metriccell{tablecost}{5.0}{2} & \metriccell{tablecost}{45.0}{\textbf{1.7}} & \metriccell{tablecost}{45.0}{\textbf{1}} \\
 & Perplexity & \metriccell{tableperplexity}{25.0}{22.4\,{\scriptsize$\pm$\,0.2}} & \metriccell{tableperplexity}{25.0}{2} & \metriccell{tableperplexity}{5.0}{22.5\,{\scriptsize$\pm$\,0.2}} & \metriccell{tableperplexity}{5.0}{3} & \metriccell{tableperplexity}{25.0}{22.4\,{\scriptsize$\pm$\,0.1}} & \metriccell{tableperplexity}{25.0}{2} & \metriccell{tableperplexity}{45.0}{\textbf{22.2\,{\scriptsize$\pm$\,0.1}}} & \metriccell{tableperplexity}{45.0}{\textbf{1}} & \metriccell{tableperplexity}{45.0}{\textbf{22.3\,{\scriptsize$\pm$\,0.1}}} & \metriccell{tableperplexity}{45.0}{\textbf{1}} & \metriccell{tablecost}{5.0}{19.6} & \metriccell{tablecost}{5.0}{2} & \metriccell{tablecost}{45.0}{\textbf{3.3}} & \metriccell{tablecost}{45.0}{\textbf{1}} & \metriccell{tablecost}{5.0}{13.4} & \metriccell{tablecost}{5.0}{2} & \metriccell{tablecost}{45.0}{\textbf{9.7}} & \metriccell{tablecost}{45.0}{\textbf{1}} & \metriccell{tablecost}{5.0}{6.0} & \metriccell{tablecost}{5.0}{2} & \metriccell{tablecost}{45.0}{\textbf{1.7}} & \metriccell{tablecost}{45.0}{\textbf{1}} \\
\hline
\multirow{2}{*}{\textsc{RTX2080}} & Latency (ms) & \metriccell{tablelatency}{5.0}{21.2\,{\scriptsize$\pm$\,1.5}} & \metriccell{tablelatency}{5.0}{2} & \metriccell{tablelatency}{5.0}{21.3\,{\scriptsize$\pm$\,1.3}} & \metriccell{tablelatency}{5.0}{2} & \metriccell{tablelatency}{5.0}{21.1\,{\scriptsize$\pm$\,1.1}} & \metriccell{tablelatency}{5.0}{2} & \metriccell{tablelatency}{45.0}{\textbf{18.9\,{\scriptsize$\pm$\,1.2}}} & \metriccell{tablelatency}{45.0}{\textbf{1}} & \metriccell{tablelatency}{5.0}{21.4\,{\scriptsize$\pm$\,1.5}} & \metriccell{tablelatency}{5.0}{2} & \metriccell{tablecost}{5.0}{19.1} & \metriccell{tablecost}{5.0}{2} & \metriccell{tablecost}{45.0}{\textbf{3.2}} & \metriccell{tablecost}{45.0}{\textbf{1}} & \metriccell{tablecost}{5.0}{13.4} & \metriccell{tablecost}{5.0}{2} & \metriccell{tablecost}{45.0}{\textbf{9.0}} & \metriccell{tablecost}{45.0}{\textbf{1}} & \metriccell{tablecost}{5.0}{6.0} & \metriccell{tablecost}{5.0}{2} & \metriccell{tablecost}{45.0}{\textbf{1.7}} & \metriccell{tablecost}{45.0}{\textbf{1}} \\
 & Perplexity & \metriccell{tableperplexity}{18.3}{22.4\,{\scriptsize$\pm$\,0.2}} & \metriccell{tableperplexity}{18.3}{3} & \metriccell{tableperplexity}{5.0}{22.5\,{\scriptsize$\pm$\,0.2}} & \metriccell{tableperplexity}{5.0}{4} & \metriccell{tableperplexity}{18.3}{22.4\,{\scriptsize$\pm$\,0.1}} & \metriccell{tableperplexity}{18.3}{3} & \metriccell{tableperplexity}{45.0}{\textbf{22.2\,{\scriptsize$\pm$\,0.1}}} & \metriccell{tableperplexity}{45.0}{\textbf{1}} & \metriccell{tableperplexity}{31.7}{22.3\,{\scriptsize$\pm$\,0.1}} & \metriccell{tableperplexity}{31.7}{2} & \metriccell{tablecost}{5.0}{19.7} & \metriccell{tablecost}{5.0}{2} & \metriccell{tablecost}{45.0}{\textbf{3.3}} & \metriccell{tablecost}{45.0}{\textbf{1}} & \metriccell{tablecost}{5.0}{13.3} & \metriccell{tablecost}{5.0}{2} & \metriccell{tablecost}{45.0}{\textbf{9.6}} & \metriccell{tablecost}{45.0}{\textbf{1}} & \metriccell{tablecost}{5.0}{6.0} & \metriccell{tablecost}{5.0}{2} & \metriccell{tablecost}{45.0}{\textbf{1.7}} & \metriccell{tablecost}{45.0}{\textbf{1}} \\
\bottomrule
\end{tabular}
\endgroup
}
\end{table*}

\subsubsection{Results}
\textit{Tuning quality.} \jev{} reaches rank 1 on two tasks, compared with three for the original \promisetune{} and four for \deepseek{} (Table~\ref{tab:region}). Its clearest improvement over the original tuner is perplexity, which falls from 22.392 to 22.266 on \textsc{A100} and 22.267 on \textsc{RTX2080}. The paired differences are $-0.126$ and $-0.125$ (95\% intervals $[-0.196,-0.056]$ and $[-0.191,-0.062]$; adjusted $p<0.02$). Both models also outrank \treeregion{} on these tasks. For efficiency, \deepseek{} leads both latency comparisons and \treeregion{} leads \textsc{Covertype} training time, indicating that the choice of guider depends on the target metric.

\begin{figure*}[t]
\centering\includegraphics[width=\textwidth]{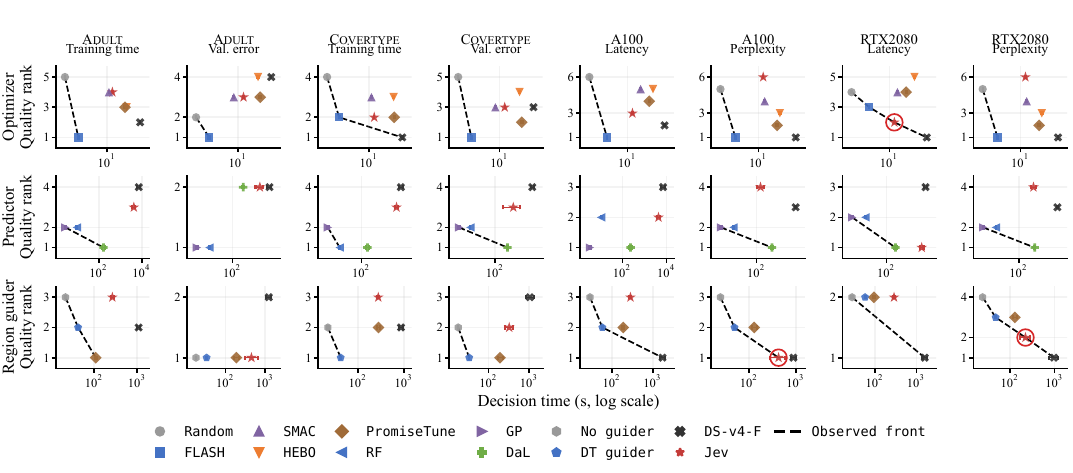}
\caption{(RQ1--RQ3) Performance-rank--decision-time trade-offs at 100 configurations. Horizontal bars show 95\% intervals. Dashed lines connect Pareto-optimal methods. Red rings mark \jev{} on the front.}
\label{fig:tradeoff}
\end{figure*}

\textit{Quality--cost trade-off.} The perplexity gains over native \promisetune{} come with longer selection time and API fees. \jev{} lies on the rank--time front in both GPU perplexity comparisons, at rank 1 on \textsc{A100} and rank 2 on \textsc{RTX2080} (Figure~\ref{fig:tradeoff}).

Among the API guiders, \jev{} costs 62--72\% less than \deepseek{}. The validation-error tasks favor \jev{} on both mean quality and fees, while the remaining six trade its lower fee against \deepseek{}'s better mean performance (Table~\ref{tab:region}).

\begin{rqanswer}
\textbf{RQ3 Answer.} \jev{}'s region guidance finds lower-perplexity configurations than native \promisetune{}, trading additional selection time and API fees for that quality gain.
\end{rqanswer}

\section{Discussion}
\label{sec:discussion}
\subsection{Choosing an Optimizer}
\jev{}'s weaker optimizer results than \flash{}'s coincide with persistent reuse of earlier proposals, which account for more than half of its proposals on average in every task. Under nearest-unseen mapping, each repetition selects the next available configuration by distance to the same target. Across these runs, \jev{}'s last 30 new configurations have worse mean performance than its first 30 after initialization. \flash{} improves over the same windows as it refits its predictor and ranks unseen configurations after each measurement.

\begin{implicationbox}
\textbf{Implication 1.} Use \flash{} as the starting choice when configuration quality is the priority, and consider \deepseek{} when only a small number of configurations can be evaluated. \jev{} lowers API fees but usually finds worse configurations than \deepseek{}.
\end{implicationbox}

\subsection{Choosing a Performance Predictor}
\jev{} selects the best candidate in its pool less often than \dal{} across all eight tasks, helping explain its weaker tuning results. For GPT perplexity, \jev{} selects the true best candidate in 7--8\% of prediction-guided steps, compared with about 42\% for \dal{}. Although both models capture the broad ordering of these candidates, \jev{} more often misorders the best-performing alternatives. Because the search evaluates the lowest predicted candidate, these errors directly divert measurements from better configurations. Reliable ordering of strong candidates is therefore central to predictor choice, complementing evidence that numerical accuracy gains must be sufficiently large to improve tuning~\cite{accuracycanlie}.
\begin{implicationbox}
\textbf{Implication 2.} Retain conventional performance predictors inside the tuner. With \dal{} as the predictor, the tuner finds configurations that rank first, including ties, in seven of eight comparisons.
\end{implicationbox}

\subsection{Choosing a Region Guider}
\jev{}'s lower perplexity than native \promisetune{} and \treeregion{} appears linked to recommendations that place at least five feed-forward width ratios in the upper half of their allowed ranges. Across both GPUs, configurations satisfying these joint width conditions are about four times as likely to fall in the benchmark's lowest-perplexity 1\% as other configurations, using the median ratio across recommendations. Over the ten-selection blocks following these recommendations, \jev{} averages 29--32\% matching configurations per run, compared with 15--18\% for native \promisetune{} and \treeregion{} at the same steps.

For speed, however, \jev{} gives weaker emphasis to shallow networks. When tuning MLP training time, it selects single-layer networks in 59--60\% of new evaluations, compared with 73--75\% for the conventional guiders, although single-layer networks account for nearly all of the fastest configurations. Its GPU latency guidance similarly focuses on attention-head counts while usually leaving depth unrestricted, despite the fastest configurations being predominantly minimum-depth. These choices spread evaluations over deeper networks, reducing the search effort directed toward fast architectures.

For validation error, the predefined interval choices give \jev{} coarser control over narrow parameter ranges than the thresholds learned by conventional guiders. This difference is relevant to \textsc{Covertype}, whose lowest-error 1\% of configurations all use weight decay below 0.0066, compared with an upper bound of 0.0944 for \textsc{Adult}'s corresponding configurations. The finest low-weight-decay interval available through our model interface ends at 0.025, so it combines \textsc{Covertype}'s narrow range with substantially stronger regularization settings. On \textsc{Covertype}, \jev{} selects 20\% of new configurations with weight decay below 0.0066, compared with 24--26\% for native \promisetune{} and \treeregion{}.

\begin{implicationbox}
\textbf{Implication 3.} \jev{} can improve configuration quality through region guidance within an existing tuner, trading additional decision time and API fees for better final configurations.
\end{implicationbox}

Future work could fine-tune open \jev{}-style decision models for these roles, building on the domain adaptation demonstrated by \texttt{SecJev} and \texttt{EO-Jev}~\cite{secjev,eojev}. For optimization, \flash{} trajectories could supervise the next configuration choice from each updated history, with a training penalty for repeating earlier proposals. For prediction, training could emphasize correctly ordering pairs of high-performing candidates, targeting the reversals that divert evaluations from better configurations. For region guidance, complete candidate regions could be ranked by the proportion of high-performing configurations they contain for the requested metric. Finer interval choices would let this training distinguish narrow effective ranges that the current coarse options combine with poorer settings.

\section{Threats to Validity}
\label{sec:threats}
\textbf{Construct validity.} Prediction accuracy and tuning utility can yield different method orderings~\cite{accuracycanlie}. We measure both on matched tasks and assess output-range sensitivity by clipping conventional predictions to the models' training-label range. Conventional predictors retain their MAPE advantage across all eight tasks. Variation across execution nodes affects decision time, so we retain node identifiers and report bootstrap intervals over the 30 original run totals to quantify observed timing variation.

\textbf{Internal validity.} Encoding, initialization, and data splits affect method comparisons~\cite{gongencoding}. We fix feature encodings, pair initial samples, and use identical disjoint prediction splits. We also record RQ3 refresh schedules, treating the native-tuner comparison as the combined effect of recommender and update frequency.

\textbf{Statistical conclusion validity.} Sampling variation can change empirical comparisons, as prior work on repeated DL evaluations illustrates~\cite{jahangirovamutation}. In our tuning study, we address variation in the sampled configurations through 30 paired seeds and report dispersion and paired intervals. Following common practice in configuration-performance research~\cite{gongdeep}, we use Scott--Knott ESD ranks and paired Wilcoxon tests with Holm adjustment. Together, these analyses show the variability, magnitude, and statistical support of the observed differences.

\textbf{External validity.} Workload and hardware changes can alter configuration performance~\cite{sempl}. Our scope includes fixed recorded workloads and one decision model and one LLM; the GPT perplexity comparisons share one quality landscape. To broaden the comparison, we study MLP training and GPT inference from two published benchmarks~\cite{lcbench,hwgpt}, covering efficiency and predictive quality in four system contexts. This diversity reveals the role-dependent outcomes in RQ1--RQ3.

\section{Related Work}
\label{sec:related}
\subsection{Configuration Tuning and Performance Learning}
ML testing examines system quality across data, learning programs, and frameworks~\cite{zhangmltesting}, including development faults~\cite{humbatovafaults}. Configuration tuning improves performance by choosing settings for a given implementation. Predictor-guided tuners use learned estimates to select configurations~\cite{flash,smac,hebojournal}, while region-guided tuners additionally recommend search regions~\cite{promisetune}. Auxiliary metrics~\cite{mmo}, adaptive weights~\cite{admmo}, and performance aspirations~\cite{aspirations} express different priorities for this search.

Within such tuners, performance learning depends on how configurations are represented and sampled~\cite{gongdeep,gongencoding}. \dal{} learns local predictors through adaptive partitions~\cite{daloriginal,daljournal}, while \texttt{CoMSA} adds uncertainty-guided sampling~\cite{comsa}. Other work adapts performance knowledge to new environments~\cite{sempl}, drift~\cite{dhda}, and software revisions~\cite{iopv}. Our study examines predictor choice within a fixed search policy, relating numerical error to the performance of selected configurations.

\subsection{Decision Models and Empirical Evaluation}
Independent evaluations examine \jev{} on classification and sequential games~\cite{jevbenchmark}, knowledge and mathematical reasoning~\cite{jev}, and tool use and social simulation~\cite{jeval}. Agent studies find varying effects on success and total cost~\cite{jevworkflow}, while ecosystem evidence documents action selection, filtering, and model/tool routing~\cite{jevwild}.

Open models offer additional implementations. \texttt{JEVal} evaluates \texttt{Kev}, \texttt{Intern-Decision}, \texttt{Laya}, and \texttt{Nimble}, and introduces \texttt{InnerJev} through reasoning-to-readout distillation~\cite{jeval}. \texttt{Jeff} adapts LLM decision heads and is evaluated for accuracy and calibration~\cite{openjev}; \texttt{LLM2Jev} studies option probabilities and fine-tuning for typed decisions~\cite{llm2jev}. We complement these model and agent evaluations with evidence on AI tuning, where successive decisions change the available observations and the best configuration reached.

\subsection{Model-Assisted Optimization}
\texttt{OptFormer} learns optimization policies and predictions from earlier tuning experiments~\cite{optformer}. \texttt{LLAMBO} uses LLMs for initialization, surrogate modeling, and candidate sampling~\cite{llambo}. Gupta et al.~\cite{llmboempirical} find that classical optimizers outperform LLM agents, while their hybrid \texttt{LLMNN} combines model guidance with nearest-neighbor sampling to achieve competitive or superior results. This supports studying model--algorithm combinations; our comparison distinguishes three decision roles in configuration tuning.

LLMs also optimize code~\cite{gongcodesurvey} through contextual prompts~\cite{mpco} and model ensembles~\cite{industrialmoa,moasurvey}. Semantic-transformation benchmarks~\cite{segabench}, efficiency benchmarks~\cite{effibench}, re-executed agent fixes~\cite{mergedmeasured}, and execution-mode migrations~\cite{graphmigration} expose performance gains and regressions. Pipeline tests~\cite{pipelinetests} and generated microbenchmarks~\cite{llm4jmh} supply measurement evidence, while \llmperf{} uses LLMs for performance-model sampling~\cite{llm4perf}. Our comparison keeps target implementations fixed and evaluates a decision model alongside an LLM and conventional methods across complete optimization, prediction, and region guidance.

\subsection{Quality--Cost Optimization}
Quality--cost optimization appears in both AI generation and agent design. \texttt{GreenStableYolo} tunes image-generation parameters and prompts~\cite{greenstableyolo}, while \texttt{GA4GC} considers agent runtime and generated-code performance~\cite{ga4gc}. \texttt{COMPAS} learns difficulty-specific quality--cost fronts~\cite{compas}, and \texttt{SkillMOO} and \texttt{ARTEMIS} evolve agent skills and components~\cite{skillmoo,evolvingexcellence}. In DNN testing, \texttt{FuzzInsight} examines how computation supports behavioral exploration~\cite{fuzzinsight}. Our study measures the cost of making configuration decisions and relates it to the performance of the configurations they produce.

\section{Conclusion}
Conventional methods generally find better configurations in direct optimization and predictor-guided search, while \jev{} helps an existing tuner find lower-perplexity configurations through region guidance at additional decision time and API fees. To improve decision models across these roles, future work could fine-tune open models on tuning histories and measured performance, teaching them to avoid repeated proposals, correctly order strong candidates, and recommend regions with a higher proportion of high-performing configurations.


\clearpage
\bibliographystyle{IEEEtran}
\bibliography{references}
\end{document}